# Improving the Quality of Persian Clinical Text with a Novel Spelling Correction System

Seyed Mohammad Sadegh Dashti[1], Seyedeh Fatemeh Dashti[2]

[1] Computer Engineering Department, Kerman Branch, Islamic Azad University, Kerman, Iran

[2] Department of Advanced Research, Bushehr University of Medical Sciences, Bushehr, Iran

Corresponding Author: Seyed Mohammad Sadegh Dashti; dashti@iauk.ac.ir

**Abstract**

**Background:** The accuracy of spelling in Electronic Health Records (EHRs) is a critical factor for efficient clinical care, research, and ensuring patient safety. The Persian language, with its abundant vocabulary and complex characteristics, poses unique challenges for real-word error correction. This research aimed to develop an innovative approach for detecting and correcting spelling errors in Persian clinical text.

**Methods:** Our strategy employs a state-of-the-art pre-trained model that has been meticulously fine-tuned specifically for the task of spelling correction in the Persian clinical domain. This model is complemented by an innovative orthographic similarity matching algorithm, PERTO, which uses visual similarity of characters for ranking correction candidates.

**Results:** The evaluation of our approach demonstrated its robustness and precision in detecting and rectifying word errors in Persian clinical text. In terms of non-word error correction, our model achieved an F1-Score of 90.0% when the PERTO algorithm was employed. For real-word error detection, our model demonstrated its highest performance, achieving an F1-Score of 90.6%. Furthermore, the model reached its highest F1-Score of 91.5% for real-word error correction when the PERTO algorithm was employed.

**Conclusions:** Despite certain limitations, our method represents a substantial advancement in the field of spelling error detection and correction for Persian clinical text. By effectively addressing the unique challenges posed by the Persian language, our approach paves the way for more accurate and efficient clinical documentation, contributing to improved patient care and safety. Future research could explore its use in other areas of the Persian medical domain, enhancing its impact and utility.

**Keywords:** real-word error; non-word error; spelling correction; contextualized embeddings; deep learning; Radiology reporting

# Abbreviations
EHR    Electronic health record
OCR    Optical character recognition
ASR    Automatic Speech Recognition
NLP    Natural Language Processing
HIS    Hospital information system
BCI    Brain-Computer Interfaces
VR    Virtual Reality
AR    Augmented Reality
EEG    Electroencephalography
MLM    Masked Language Model

## 1. Introduction

Spelling correction is a vital task in all text processing environments, with its importance amplified for languages with intricate morphology and syntax, such as Persian. This significance is further heightened in the realm of clinical text, where precise documentation is a cornerstone for effective patient care, research, and ensuring patient safety The written text of medical findings remains the essential source of information for clinical decision making. Clinicians prefer to write unstructured text rather than filling out structured forms when they document the progress notes, due to time and efficiency constraints [1]. The quality and safety of health care depend on the accuracy of clinical documentation [2]. However, misspellings often occur in clinical texts because they are written under time pressure [3].

The process of spelling correction primarily tackles two types of errors: non-word errors, which are nonsensical words not found within a dictionary, and real-word errors, that are correctly spelled words but utilized inappropriately in context. These errors can stem from various sources including typographical mistakes, confusion between similar sounding or meaning words [4], incorrect replacements by automated systems like AutoCorrect features [5], and misinterpretation of input by ASR and OCR systems [6-9].

The Persian language, with its rich vocabulary and complex properties, presents unique challenges for real-word error correction. Features unique to Persian such as homophony (words that are pronounced identically yet carry distinct meanings), polysemy (words with multiple meanings), heterography (words that share identical spelling but their meanings vary based on how they are pronounced), and word boundary issues contribute to this complexity.

    Despite these challenges, numerous efforts have been made to develop both statistical and rule-based approaches for identifying and rectifying both classes of errors in the general Persian text domain; however, the work in the Persian medical domain and specifically the Persian clinical text is very limited. Moreover, these methods have attained only limited success. In this study, we introduce an innovative method to detect and correct word errors in Persian clinical text, aiming to significantly improve the accuracy and reliability of healthcare documentation. Our key contributions include:

- Language Representation Model: We showcase a pre-trained language representation model that has undergone meticulous fine-tuning, specifically for the task of spelling correction in the Persian clinical domain.

- PERTO Algorithm: We introduce an innovative orthographic similarity matching algorithm that leverages the visual resemblance of characters to prioritize correction candidates.

We utilize the F1-score metric to evaluate and contrast our methodology with established approaches for detecting and rectifying both non-word and real-word errors within the context of Persian clinical text.

The rest of this paper is structured as follows: We commence with a review of prior research in the field. Following this, we delve into the challenges faced in Persian language text processing. Subsequently, we outline our proposed approach. Evaluation and experiment results are then presented and discussed. In the final segment, we summarize our findings.

## 2. Related Works

Automatic word error correction is a crucial component in NLP systems, particularly in the context of EHR and clinical reports. Early techniques were based on edit distance and phonetic algorithms [10-13]. The incorporation of context information has been demonstrated to be effective in boosting the efficiency of auto-correction systems [14]. Contextual measures like semantic distance and noisy channel models based on N-grams have been employed across numerous NLP applications [4, 5, 15-17]. A novel approach was also developed to correct multiple context-sensitive errors in excessively noisy situations [18]. Dashti developed a model that addressed the identification and automatic correction of context-sensitive errors in cases where more than one error existed in a given word sequence [19].

Cutting-edge methods in NLP systems utilize context information through neural word or sense embeddings for spelling correction [20]. Pretrained contextual embeddings have been used to detect and rectify context-sensitive errors [21]. The issue of spelling correction has been addressed using deep learning techniques for various languages in recent years. For example, a study in 2020 proposed a deep learning method to correct context-sensitive spelling errors in English documents. [22]. Another work developed a BERT-Based model for the same purpose [23]. NeuSpell is a user-friendly neural spelling correction toolkit that offers a variety of pre-trained models [24]. SpellBERT is a lightweight pre-trained model for Chinese spelling check [25]. A disentangled phonetic representation approach for Chinese spelling correction was proposed [26]. Other approaches for Chinese spelling correction utilized phonetic pre-training [27]. An innovative approach was devised specifically for the purpose of contextual spelling correction within comprehensive speech recognition systems [28]. A dual-function framework for detecting and correcting spelling errors in Chinese was proposed [29]. Liu and colleagues proposed a method, known as CRASpell, which is resilient to contextual typos and has been developed to enhance the process of correcting spelling errors in Chinese [30]. AraSpell is an Arabic spelling correction approach that utilized a Transformer model to understand the connections between words and their typographical errors in Arabic [31].

In the realm of healthcare, the application of spelling correction techniques has been instrumental in expanding acronyms and abbreviations, truncating, and rectifying misspellings. It has been observed that such instances constitute up to 30% of clinical content [32]. In the last twenty years, a significant amount of research has been conducted on spelling correction methods specifically designed for clinical texts [1]. The majority of these studies have primarily focused on EHR [33], while a few have explored consumer-generated texts in healthcare [34, 35].

Several noteworthy contributions in this field include the French clinical record spell checker introduced by Ruch and colleagues, which boasts a correction rate of up to 95% [36].

Siklósi and his associates devised a system that is aware of context for Hungarian clinical text, which is grounded on statistical machine translation, and it attained an accuracy rate of 87.23% [37]. Grigonyte and her research team introduced a system tailored for Swedish clinical text, achieving a precision of 83.9% and a recall rate of 76.2%. [38].

Zhou and colleagues leveraged the Google spell checker to develop a system capable of accurately correcting 86% of typographical and linguistic inaccuracies found in routine medical terminologies [35]. Another study deliberated on a spelling correction system that was referenced in reports concerning the safety of vaccines, with recall and precision rates of 74% and 47%, respectively [39]. Wong and his team have designed a system that operates in real-time to rectify spelling errors in clinical reports, achieving an accuracy of 88.73%. This system leverages the power of semantic and statistical analysis applied to web data for the purpose of automatic correction [1]. Doan and his research team presented a system, specifically designed for the rectification of misspellings in drug names. This system, which is based on the Aspell algorithm, reported a commendable precision rate of 80% [40].

Among the recent contributions is an article by Lai and colleagues proposing a system for automatic spelling correction in medical texts, employing a noisy channel model to achieve significant accuracy [41]. Similarly, unsupervised, context-aware models have shown promise in correcting spelling errors in English and Dutch clinical unstructured texts [42, 43].

While these advancements have significantly improved spelling correction across languages and domains, recent innovations in BCIs, eye-tracking, VR/AR, and non-invasive EEG technologies open new avenues for further enhancing human-computer interaction and the accuracy of medical documentation [44-47]. These technologies, through their unique capabilities to interact directly with the user's cognitive states and attention, offer potential solutions to some of the inherent limitations of current NLP systems in understanding and correcting complex, context-sensitive errors in clinical texts. As the field continues to evolve, integrating these cutting-edge technologies into spelling correction tools for medical documentation could revolutionize the way healthcare professionals interact with digital text, making the process more efficient, accurate, and tailored to their specific needs.

In addition, the emergence and application of Optical technology in the healthcare sector over the past twenty years has led to the creation of several systems designed to detect and correct OCR errors automatically. A reference to one such system can be found in [48]; this system identifies and rectifies typographical errors in French clinical documents. In a newer study, Tran and colleagues suggested a model for spelling correction in clinical text that is sensitive to context. [49].

Despite the complexities inherent in the Persian language, substantial progress has been made in the field of spelling correction. The strategies employed range from statistical or rule-based methods to more contemporary systems, such as the Vafa spellchecker, which is capable of detecting a wide variety of errors. Mosavi and Miangah have addressed spelling issues in the Persian language using N-grams, a monolingual corpus, and a measure of string distance [50-58]. Within these methodologies, one focuses on correcting typographical errors in clinical text, utilizing a four-gram language model. Consequently, the need for a Persian spell-checking tool in specialized domains, such as healthcare, is clear.

Given the variety of methodologies and their targeted applications in spelling correction, we provide Table 1 below to efficiently summarize the key contributions within the medical domain and Persian language spelling correction models. This comparative analysis not only illuminates the range of strategies employed to address spelling correction challenges across diverse languages and contexts but also underlines the distinctive features of each method. In doing so, it enhances our understanding of the current research landscape in this field, spotlighting the innovative approaches and shedding light on the potential avenues for future exploration.

Table 1. Comparative Analysis of Spelling Correction Models Across Languages with a Focus on the Medical Domain

| Model | Objective | Language | Method Category | Dataset Description | Application Domain |
|---|---|---|---|---|---|
| **Fivez et al., 2017** | Unsupervised and context-sensitive spelling correction | English, Dutch | Deep Learning, Statistical/ Rule-Based | Public corpora (MIMIC-III, Health forums), Proprietary clinical records | Medical/Clinical |
| **Lai et al., 2015** | Automated error detection and correction | English | Statistical/ Rule-Based | Clinical reports, Clinical notes | Medical/Clinical |
| **Kilicoglu et al., 2015** | Spelling correction in consumer health questions | English | Statistical/ Rule-Based | Consumer health questions | Medical/Clinical |
| **Hussain and Qamar, 2016** | Improving text mining for better information retrieval | English | Machine Learning | Medical documents | Medical/Clinical |
| **Yazdani et al., 2019** | Automated misspelling detection and correction in Persian clinical texts | Persian | Statistical/ Rule-Based | Persian radiology and ultrasound reports | Medical/Clinical |
| **Dastgheib et al., 2016** | Semantic-based spelling correction for Persian | Persian | Statistical/ Rule-Based | Persian parallel corpus | General Text |
| **Faili et al., 2014** | Automatic detection of spelling, grammatical, and real-word errors in Persian texts | Persian | Statistical/ Rule-Based | Persian digital texts | General Text |

**2.1 Persian Spelling Challenges**

Persian, alternatively referred to as Farsi, belongs to the Indo-Iranian subgroup of the Indo-European family of languages. It holds official language status in countries such as Iran, Tajikistan, and Afghanistan. Over time, Persian has incorporated elements from other languages such as Arabic, thereby enriching its vocabulary. Despite these influences, the fundamental structure of the language has largely remained intact for centuries [55, 59].

While Persian is a vibrant and expressive language, it presents several challenges for language processing:

1. **Character Ambiguity**: Persian characters like "ی" and "ي" are often used interchangeably but represent different sounds [60].
2. **Rich Morphology**: New words can be created by adding prefixes and suffixes to a base word, like "دست" (hand) to "دست‌ها" (hands) [61].
3. **Orthography**: Persian involves a combination of spaces and semi-spaces, which can lead to inconsistencies [62].
4. **Co-articulation**: The pronunciation of a consonant like "ب" can be affected by the subsequent vowel [63].
5. **Dialectal Variation**: Persian has several standard varieties such as Farsi, Dari, and Tajik [64].
6. **Cultural Factors**: The phenomenon of persianization can shape the way Persian is used and interpreted.
7. **Lack of Resources**: Often, Persian is classified as a language with limited resources, given the scarcity of accessible data and tools for Natural Language Processing [61].
8. **Free Word Order**: Persian allows for the rearrangement of words within a sentence without significantly altering its meaning [65].
9. **Homophony**: Different words have identical pronunciation but different meanings, like ("گذار" /gʊzɑr/ 'transition')[1] and ("گزار" /gʊzɑr/ 'predicate') [66].
10. **Diacritics**: They are frequently left out in writing, leading to ambiguity in word recognition [67].
11. **Rapidly Changing Vocabulary**: Persian's vocabulary is rapidly evolving due to factors such as technology, globalization [68].
12. **Lack of standardization**: There isn't a single standard for Persian text, which can complicate the development of language processing models capable of handling a variety of dialects and styles [69].

A significant issue is the treatment of internal word boundaries, often represented by a zero-width non-joiner space or "pseudo-space". Ignoring these can lead to text processing errors. Pre-processing steps can help resolve these issues by correcting pseudo and white spaces according to internal word boundaries and addressing tokenization problems.

These challenges highlight the need for robust computational models and resources that can handle the intricacies of the Persian language while ensuring accurate language processing.

**3. Material and Methods**

---

[1] All pronunciations have been provided in International Phonetic Alphabet (IPA)

Our methodology detects and corrects two categories of mistakes in Persian clinical text: Non-word and Real-word errors. The architecture of the proposed system is depicted in Figure 1. The system design is composed of five distinct modules that communicate via a databus.

The INPUT module accepts raw test corpora. The pre-processing component normalizes the text and addresses word boundary issues. The contextual analyzer module assesses the contextual similarity within desired word sequences.

For error detection, we implement a dictionary reference technique to pinpoint non-word errors and use contextual similarity matching to detect real-word errors. The error correction module rectifies both classes of errors using context information from a fine-tuned contextual embeddings model, in conjunction with orthographic and edit-distance similarity measures.

The corrected corpora or word sequence is then delivered through the OUTPUT module.

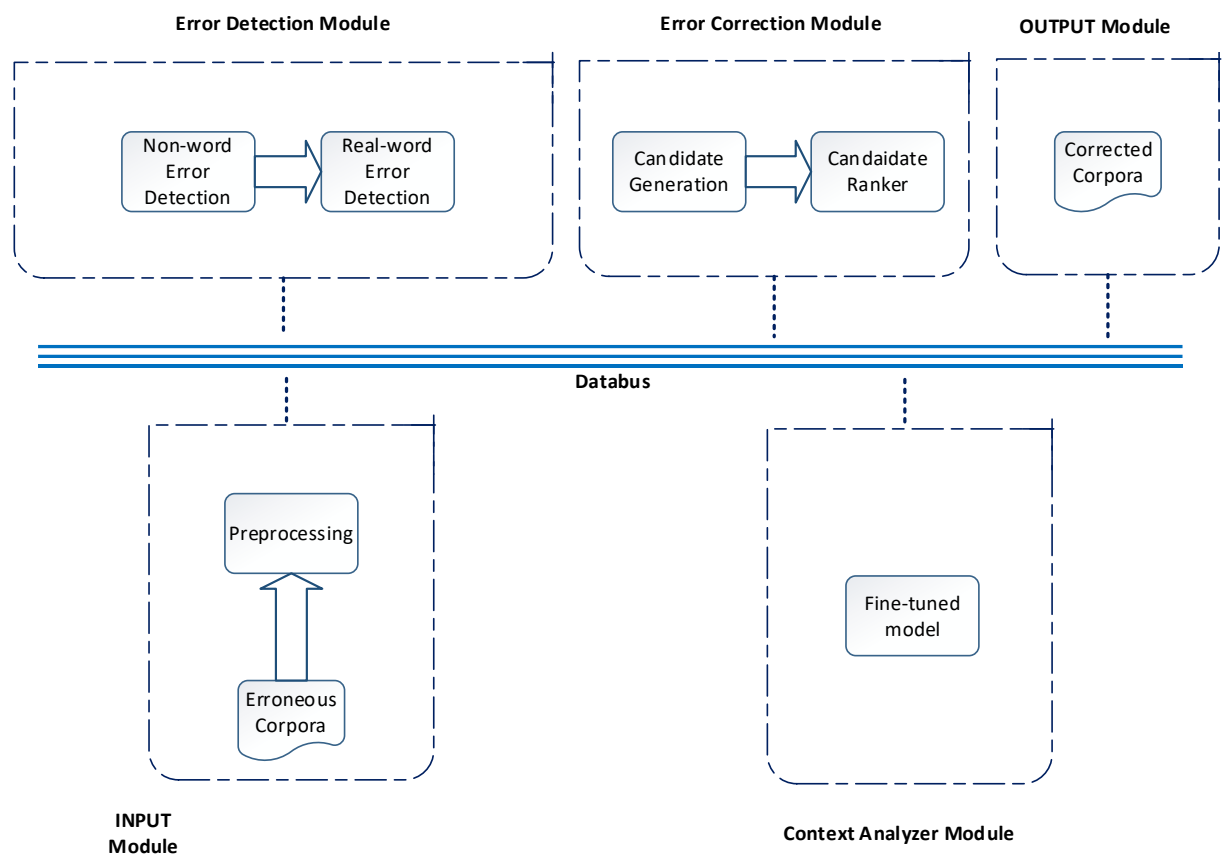

Figure 1. Architecture of the Proposed System for Detecting and Correcting Persian Word Errors.

## 3.1 Pre-processing step

Text pre-processing is a crucial step in numerous NLP applications, which includes the segmentation of sentences, tokenization, normalization, and the removal of stop-words. The segmentation of sentences involves determining the boundaries of a sentence, usually marked by punctuation such as full stops, exclamation marks, or question marks. Tokenization is the process of decomposing a sentence into a set of terms that capture the sentence's meaning and are utilized for feature extraction. Normalization is the procedure of converting text into its standard forms and is particularly important in NLP applications for Persian, as it is for many other languages. A key task in normalizing Persian text is the conversion of pseudo and white spaces into regular forms, replacing whitespaces with zero-width non-joiners when necessary.

For example, ('می شود' /mi ʃævæd/ 'is becoming') is replaced with ('میشود' / miʃævæd / 'is becoming'). Persian and Arabic have numerous similarities, and certain Persian alphabets are frequently incorrectly written using Arabic versions. It is often advantageous for researchers to normalize these discrepancies by substituting Arabic characters (ي 'Y' /j/; ک 'k' /k/; ه 'h' /h/) with their corresponding Persian forms. For instance, ('براي' /bæray/ 'for') is transformed to ('برای' /bæray/ 'for'). Normalization also includes removing diacritics from Persian words; e.g., ('ذرّه' /zærre/ 'particle) is changed to ('ذره' /zære/ 'particle). Additionally, Kashida(s) are removed from words; for instance, ('بـــاند' /band/ 'band') is transformed to ('باند' /band/ 'band').

In order to accomplish the goal of normalization, a dictionary named Dehkhoda, which includes the correct typographic form of all Persian words, is utilized to determine the standard form of words that have multiple shapes [70].

### 3.2 Damerau-Levenshtein distance and candidate generation

Our methodology employs the Damerau-Levenshtein distance metric to generate potential rectifications for both non-word and real-word errors. [11]. This measure considers insertion, deletion, substitution, and transposition of characters. For instance, the measure of Damerau-Levenshtein distance between "KC" and "CKE" equals 2. It's found that around 80% of human-generated spelling errors involve these four error types [71]. Studies indicate that context-sensitive error constitute approximately 25% to 40% of all typographical errors in English documents. [72, 73].

Our model utilizes an extensive dictionary to pinpoint misspellings. This dictionary is bifurcated into two segments: general and specialized terms. For the general segment, we employ the Vafa spell-checker dictionary, a highly respected spell checker for the Persian language. This dictionary encompasses 1,095,959 terms, all of which are general terms, but it excludes specialized medical terminology. In this research, we utilized the texts we trained to formulate a custom dictionary. This dictionary integrates specialized terminology found in breast ultrasonography, head and neck ultrasonography, and abdominal and pelvic ultrasonography texts. It was further enriched with translations from the Radiological Sciences Dictionary by David J Dowsett to pinpoint misspellings of specialized terms [74]. This dictionary comprises 10,332 terms, all of which are specialized terms in the field of breast ultrasound, head and neck ultrasound, and abdominal and pelvic ultrasound. However, this specialized dictionary does not encompass general terms.

To circumvent duplication of specialized terms, we juxtaposed our comprehensive dictionary with the Radiological Sciences Dictionary using a custom software developed by the researchers of this study. This ensured that no term was included more than once in the dictionary, as some terms might be present in both dictionaries.

Upon our analysis of the test data, we concluded that an edit distance of up to 2 between the candidate corrections and error would be ideal. With an edit distance set to one, an average of three candidates are generated as potential replacements for a target context word. However, when the edit distance is increased to 2, the average number of generated candidates rises to 15. Correspondingly, the computation time also increases. We ensure that the generated candidates are validated against the reference lexicon.

### 3.3. Contextual embeddings

Word embeddings, which analyze vast amounts of text data to encapsulate word meanings into low-dimensional vectors [75, 76], retain valuable syntactic and semantic information [77] and are advantageous for numerous NLP applications [78]. However, they grapple with the issue of meaning conflation deficiency, which is the inability to differentiate between multiple meanings of a word.

To tackle this, cutting-edge approaches represent specific word senses, referred to as contextual embeddings or sense representation. Context-sensitive word embedding techniques such as ELMo consider the context of the input sequence [64]. There exist two main strategies for pre-training language representation model: feature-oriented methods and fine-tuning methods [79]. Fine-tuning techniques train a language model utilizing large datasets of unlabeled plain texts. The parameters of these models are later fine-tuned using data that is pertinent to the task at hand [79-81]. However, pre-training an efficient language model demands substantial data and computational resources [82-85]. Models that are multilingual have been formulated for languages that share morphological and syntactic structures. However, languages that do not use the Latin script significantly deviate from those that do, thereby requiring an approach that is specific to each language [86]. This challenge is also common in the Persian language. Although some multilingual models encompass Persian, their performance may not match that of monolingual models, which are specifically trained on a language-specific lexicon with more extensive volumes of Persian text data. As far as we are aware, ParsBert [87] and SinaBERT [88] are the sole efforts to pre-train a Bidirectional Encoder Representation Transformer (BERT) model explicitly for the Persian language.

### 3.3.1 Pre-trained Language Representation Model

Persian is often recognized as an under-resourced language. Despite the existence of language models that support Persian, only two, namely ParsBert [87] and SinaBERT [88], have been pre-trained on large Persian corpora. ParsBERT was pre-trained on data from the general domain, which includes a substantial amount of informal documents such as user reviews and comments, many of which contain misspelled words.

Conversely, SinaBERT was pre-trained on unprocessed text from the overarching medical field. The data for SinaBERT was compiled from a diverse set of sources such as websites that provide health and medical news, websites that disseminate scientific information about health, nutrition, lifestyle, and more, journals (encompassing both abstracts and complete papers) and conference proceedings, scholarly written materials, medical reference books and dissertations, online forums centered around health, medical and health-related Instagram pages, along with medical channels and groups on Telegram.

The data primarily consisted of general medical domain data, a portion of which was informal and contained misspellings. These factors make these pre-trained models unsuitable for Persian clinical domain spelling correction tasks. The lack of an efficient language model in this domain poses a considerable hurdle. In the subsequent section, we will explore our Persian Clinical Corpus and the procedure of pre-training our language representation model.

### 3.3.2  Data

While numerous formal general domain Persian medical texts are freely accessible, they may not be ideal for spelling correction in clinical texts. Conversely, Persian clinical texts are not widely available to the public. Nevertheless, the use of Persian clinical text is essential for pre-training a language representation model specifically for spelling correction in Persian clinical text. Consequently, we assembled a substantial collection of Persian Clinical texts to train an effective model for spelling correction in Persian.

Our data comprises a total of 78,643 ultrasonography reports, which were obtained from three distinct datasets. These datasets were generously provided by the Department of Imaging's HIS at Tehran's Imam Khomeini Hospital. For a detailed breakdown of these datasets, please refer to Table 2.

Table 2. Details of the Datasets

| Dataset Source | Number of Reports | Number of Words | Number of Sentences | Average Length of Reports |
|---|---|---|---|---|
| Jan 2011–Feb 2015 | 22,504 | 7,538,840 | 396,781 | 335 |
| Mar 2015–Jul 2018 | 15,888 | 4,782,288 | 239,114 | 301 |
| Aug 2018–Jun 2023 | 40,251 | 14,007,348 | 1,253,736 | 348 |
| **Total** | **78,643** | **26,328,476** | **1,889,631** | **336** |

Each dataset comprised three different types of medical reports: breast ultrasonography, head and neck ultrasonography, and abdominal and pelvic ultrasound reports. The first dataset, spanning from January 2011 to February 2015, included 22,504 reports with a total of 7,538,840 words. The average report length in this dataset was 335 words. The second dataset contained 15,888 reports and 4,782,288 words, encompassing all texts entered by medical typists from March 2015 to July 2018. The average length of sonography reports in this dataset was 301 words. The third dataset, which covers the period from August 2018 to June 2023, comprises 40,251 reports and a total of 14,007,348 words. All of these reports were inputted by medical typists. The average word count for the sonography reports in this dataset is 348 words. Upon analyzing the corpus, we found that 1.2% of the words in the corpora represent instances of errors, which can be classified into two types: non-word errors and real-word errors. Further scrutiny revealed that out of this 1.2% segment, non-word errors constitute 1%, while the remaining 0.2% are real-word errors.

We employed a random selection process to ensure a fair representation of the entire corpora in both the testing and training datasets. Specifically, 10% of the sentences from the corpora, amounting to 188,963 sentences, were randomly chosen for testing and evaluation. The remaining 90% of the sentences, which equates to 1,700,668 sentences, were allocated for the fine-tuning and pre-training of the model. Of these, 10% were used for fine-tuning and the rest, 90%, for pre-training. This process encompassed several steps including normalization, pre-processing, and the removal of punctuation marks, tags, and so forth. In addition, we addressed both real-word and non-word errors present in the training corpus. This meticulous approach ensures the robustness and accuracy of our model.

### 3.3.3 Model Architecture

The structure of our suggested model is founded on the original **BERT$_{BASE}$** setup, which comprises 12 hidden layers, 12 attention heads, 768 hidden sizes, and a total of 110M parameters. Our model is designed to handle a maximum token capacity of 512. The architecture of the model is depicted in Figure 2. BERT's success is often attributed to its MLM pre-training task, where it randomly masks or replaces tokens before predicting the original tokens [80]. This feature makes BERT particularly suitable for a spelling checker, as it interprets the masked and altered tokens as misspellings. In the embedding layer of BERT, each input token, denoted as $\mathbf{T_i}$, is indexed to its corresponding embedding representation, $\mathbf{ER_i}$. This $\mathbf{ER_i}$ is then forwarded to BERT's encoder layers to obtain the subsequent representation, $\mathbf{HR_i}$.

$$ER_i = \text{BERT} - \text{Embedding}(T_i) \quad (1)$$
$$HR_i = \text{BERT} - \text{Encoder}(ER_i) \quad (2)$$

In this context, both $ER_i$ and $HR_i$ belong to the real number space $R^{1*d}$, where $d$ represents the hidden dimension. Subsequently, the similarities between $HR_i$ and all token embeddings are calculated to predict the distribution of $Y_i$ over the existing vocabulary.

$$Y_i = \text{Softmax}(HR_i, \boldsymbol{E}^T) \quad (3)$$

where $\boldsymbol{E} \in R^{V*d}$ and $Y_i \in R^{1*V}$; here $V$ signifies the size of the vocabulary and $\boldsymbol{E}$ represents the BERT embedding layer. The $i$th row of $\boldsymbol{E}$ aligns with $ER_i$ in accordance with Equation 1. The ultimate rectification outcome for $T_i$ is the $T_k$ token, whose corresponding $ER_k$ exhibits the greatest similarity to $HR_i$.

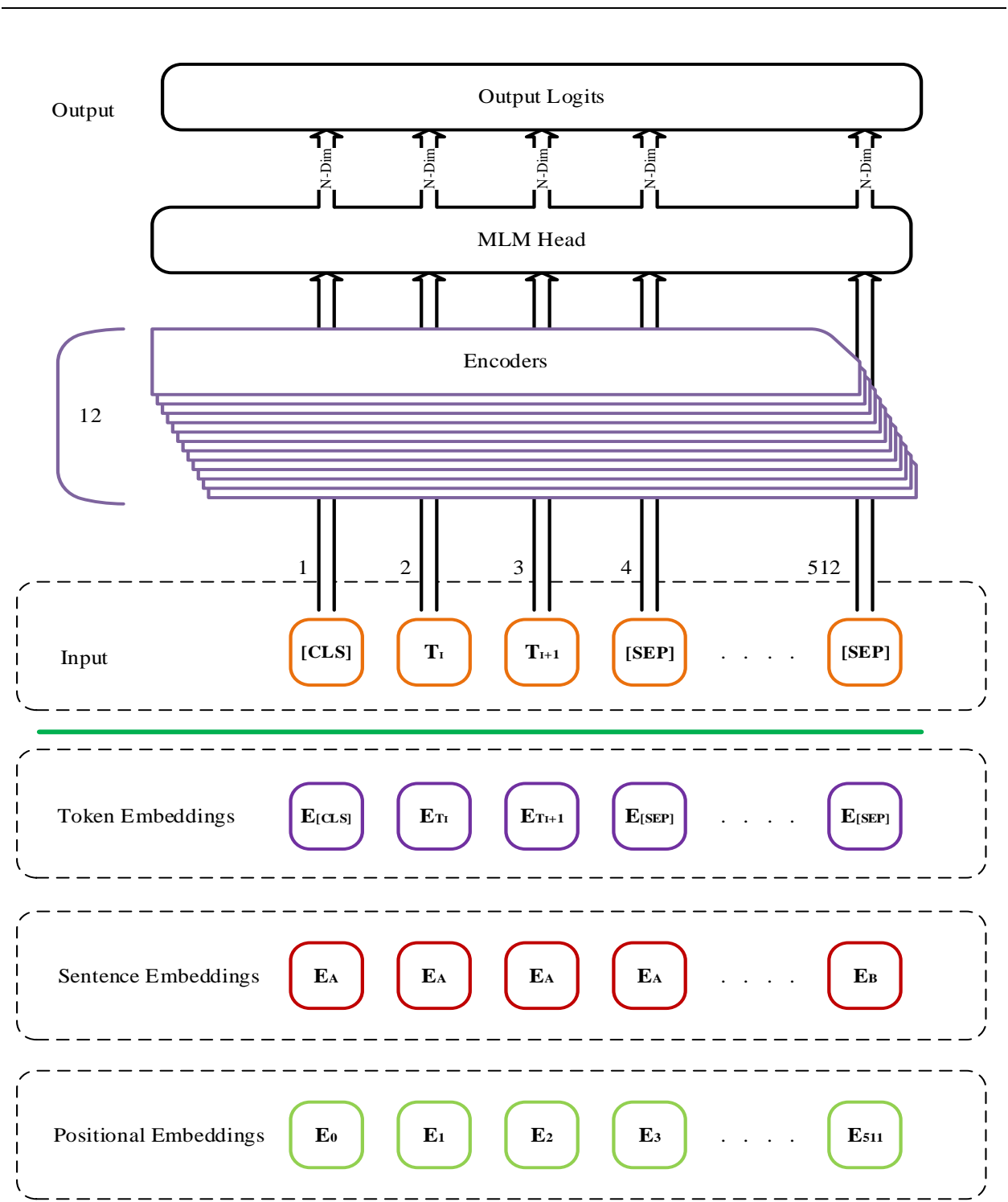

Figure 2. Architecture of Pre-trained Language Representation Model for Persian Clinical Text Spelling Correction.

### 3.4.5 Fine-tuning for Spelling Correction Task

We fine-tuned the pre-trained model specifically for the task of spelling correction in Persian clinical text, aiming to achieve optimal performance. For this fine-tuning process, we utilized 10% of the reserved sentences from the training corpus, amounting to 170,066 sentences. Each input to the model was a single sentence ending with a full stop, as our primary focus was on training the model for spelling correction. Upon examining the test set, we found that many sentences were short, and masking a few tokens would significantly reduce the context. Consequently, we excluded sentences with fewer than 20 words from the corpus. In the end, we selected 122,162 sentences, each with a minimum length of 20 words. However, since the input was a list of sentences that couldn't be directly fed into the model, we tokenized the text. The objective of the error correction task is to predict target or masked words by gaining context from adjacent words. Essentially, the model tries to reconstruct the original sentence from the masked sentence received in the input at the output. Therefore, the target labels are the actual input_ids of the tokenizer.

In the original **BERT$_{BASE}$** model, 15% of the input tokens were masked, with 80% replaced with [mask] tokens, 10% replaced with random tokens, and the remaining 10% left unchanged. However, in our fine-tuning task, we only replaced 15% of the input tokens with [mask], except for special ones; we did not use [mask] tokens to replace [SEP] and [CLS] tokens. We also avoided the random replacement of tokens to achieve better results. We used TensorFlow [89] for training with Keras [90]. Additionally, we used the Adam optimizer with a learning rate of 1E-4. The batch size was 32 and each model was run for 4 epochs.

### 3.5 PERTO Algorithm

We have designed an algorithm called PERTO, which stands for Persian Orthography Matching. This algorithm ranks the most likely candidate words derived from the output of a pre-trained model, based on shape similarity. In this algorithm, every character in the Persian script is given a distinct code. Characters that share similar forms or glyphs are classified under the same code, enabling words with similar shape characters to be identified, even if there are slight spelling variations. Our pioneering hybrid model classifies characters with the same shapes into identical groups, as depicted in Table 3.

Table 3. PERTO Code for Persian Language Alphabet

| Set Number | Hash Code | Characters with identical pronunciation |
|---|---|---|
| 1 | 0 | ا آ |
| 2 | 1 | ب پ ت ث ف |
| 3 | 2 | ج چ ح خ |
| 4 | 3 | د ذ |
| 5 | 4 | ر ز ژ |
| 6 | 5 | س ش ص ض |
| 7 | 6 | ط ظ |
| 8 | 7 | ع غ |
| 9 | 8 | ق ن ل |
| 10 | 9 | ک گ |
| 11 | A | م |
| 12 | B | و |
| 13 | C | ه |
| 14 | D | ی |

In order to identify shape similarity in Persian, a PERTO code is generated for the incorrectly spelled word. This code is subsequently matched with the PERTO codes of all potential words generated via edit distance. Our model distinctively merges PERTO with a contextual score ranking system. PERTO is solely utilized for substitution errors. In cases of insertion or deletion type errors, where the PERTO codes of all potential words do not correspond to the PERTO code of the misspelled word, our model depends entirely on contextual scores derived from the pre-trained model. Pseudocode1 outlines the implementation details of the PERTO algorithm.

To illustrate the PERTO code generation process, let us consider the word "پرگاز," which translates to "a stomach full of gas" in English. The generation of the PERTO code for this word, as per the method outlined in Pseudocode1, is as follows:

1) We begin with the first character on the right side of the word and find its hash code from Table 3. The code for "پ" is 1, which we store in an empty string.

2) Moving one unit to the left, we retrieve the hash code for the character "ر," which is 4, and add this digit to the string.

3) This process continues for each character in the word until no characters are left.

4) For "گاز," the respective codes are "9," "0" and "4," following the same lookup and concatenation procedure.

5) In the end, we obtain the PERTO code "14904" for the given word, which has the same length as the original word.

Pseudocode1: PERTO Code Generation Algorithm

| **Input:** | Word $W_i$ to be processed<br>Hashtable $H$ mapping Persian characters to PERTO codes (as per Table 3) |
|---|---|
| **Output:** | PERTO code for the input word $W_i$ |
| **Start** | |
| **1** | Initialize an empty string $PERTO\_code$; |
| **2** | For each character $C$ in $W_i$ do:<br>    Look up the character $C$ in Hashtable $H$ to find its $PERTO\_code$;<br>    Append the Hash code of $C$ to the string PERTO_Code; |
| **3** | Return $PERTO\_code$; |
| **End** | |

In the appraisal segment of our research, we will meticulously scrutinize the impact of the PERTO algorithm on the accuracy of spelling rectification within the healthcare sector. Through a comprehensive examination of the outcomes, our aim is to measure the effectiveness of this algorithm in enhancing the accuracy of spelling rectification, particularly designed for Persian medical text. This endeavor will provide valuable insights into the potential applications and benefits of the PERTO algorithm in real-world scenarios.

### 3.6  Error Detection Module

The error detection module utilizes two separate strategies based on the nature of the error being identified. For non-word errors, a lexical lookup approach is employed, while real-word errors are addressed through contextual analysis. The initial step in error detection, irrespective of the error type, involves boundary detection and token identification. Upon receiving an input sentence S, the model first demarcates the start and end of the sentence with Beginning of Sentence ($BoS$) and End of Sentence ($EoS$) markers, respectively, markers respectively, and approximates the word count in the sentence:

$$< BoS > W_i \ W_{i+1} \ W_{i+2} \ ... \ W_n < EoS >$$

It's crucial to note that the word count corresponds to the maximum number of iterations the model will undertake to identify an error in the sentence.

### 3.6.1  Non-word Error Detection

Spell checkers predominantly employ the lexical lookup method to detect spelling errors. This technique involves comparing each word in the input sentence with a reference dictionary in real-time, which is usually built using a hash table. Beginning with the $BoS$ marker, the model scrutinizes every token in the sentence for its correctness based on its sequence. This process continues until the $EoS$ marker is reached. However, if a word is identified as misspelled, the error detection cycle halts and the error correction phase commences. Here's an illustration of non-word error detection:

<div dir="rtl">در محل بررسی مایغ [مایع] مشاهده نشد.</div>

No fluid was detected in the area
that was examined.

In the given example, the word intended to be typed was ("مایع" /mɑye / 'fluid'), but it was mistakenly typed as 'مایغ'. This error is due to a substitution operation and is a single unit of distance away from the correct word. The model was successful in promptly identifying this error.

### 3.6.2 Real-word error detection

In this study, we employ contextual analysis for the detection of real-word errors. Traditional statistical models relied on n-gram language models to examine the frequency of a word's occurrence and assess the word's context by considering the frequency of the word appearing with "*n*" preceding terms. However, contemporary approaches use neural embeddings to evaluate the semantic fit of words within a given sentence. In our proposed methodology, we utilize the mask feature and leverage contextual scores derived from the fine-tuned bidirectional language model to detect and correct word errors. The process of real-word error detection is explained as follows:

1) The model begins with the *BoS* marker and attempts to encode each word as a masked word, starting with the first word.
2) A list of potential replacements for the masked word is derived from the output of the pre-trained model.
3) Based on the candidate generation scenario, replacement candidates are generated within edit-distances of 1 and 2 from the masked word.
4) The list of candidates, along with the original token, is cross-verified against the pre-trained model's output for the masked token.
5) If a candidate demonstrates a probability value that surpasses that of the masked word, the initial word is considered erroneous, thus bringing the procedure to a close.
6) However, if no error is detected, the model shifts one unit to the left, and the same steps are reiterated for all words within the sentence until the *EoS* marker is encountered.

Therefore, the moment an error is identified, the correction process is initiated immediately; subsequently, the model advances to the next sentence. Pseudocode2 offers an in-depth exploration of the Real-word error detection process.

Pseudocode2: Real-Word Error Detection Algorithm

| | |
|---|---|
| **Input:** | Sentence $S$ containing words $W_0, W_1, ..., W_n$ |
| | Language Representation Model *LRM* for probability evaluation |
| **Output:** | Sentence $S$ with identified real-word error corrected |

| Start | |
|---|---|
| **1** | For each word $W_i$ in Sentence $S_i$ do: |
| |     Mask $W_i$ and generate candidate words $C_0, C_1, ..., C_n$ from $LRM$; |
| |     Calculate probability scores for all the $C_i s$ AND $W_i$ using $LRM$; |
| |     If a candidate word $Probability(C_i) > Probability\ (W_i)$: |
| |         Mark $W_i$ as an error |
| |         break; |
| |         proceed to the Sentence $S_{i+1}$; |
| |     Else: |
| |         Keep $W_i$ unchanged in the sentence; |
| |         Move to the $W_{i+1}$ in $S_i$; |
| |         If $EOS$ is reached: |
| |         break; |
| **End** |         proceed to the Sentence $S_{i+1}$; |

Here's an illustration of successful real-world error detection:

[masked token]
در سمت چپ توده (اینترارکتال) [اینتراداکتال] دیده شد.

An intrarectal mass was seen on the left side.

In the given example, the term ( "اینتراركتال" /intrarectal/ 'intrarectal' ) is identified as a real-word error. The word that the user intended to type was ("اینتراداکتال" /intraductal/ 'intraductal' ). Initially, the model encodes the masked token and feeds it into the pre-trained model, which subsequently generates a list of contextually appropriate tokens. Following this, a roster of potential replacement candidates is created using the Damerau-Levenshtein distance measure. In this instance, the edit-distance is 2. The model then juxtaposes the context similarity score of each replacement candidate with the output list derived from the pre-trained model. Table 4 showcases the context similarity scores of the top two replacement candidates.

Table 4. Contextual Scores of the Top Five Replacement Candidates

| # | Replacement candidate | Contextual Score |
|---|---|---|
| 1 | اینتراداکتال | 0.630 |
| 2 | اینتراركتال | 0.034 |

### 3.7 Error Correction Module
The error correction phase is initiated when an error is identified in the input. In this stage, we devise a ranking algorithm that primarily relies on the contextual scores obtained from the fine-tuned pre-trained model and the corresponding PERTO codes between potential candidates and the errors.

### 3.7.1 Non-word Error Correction Process
In the non-word error correction process, the following steps are undertaken:

1) The model initially employs the Damerau-Levenshtein edit distance measure to generate a set of replacement candidates within 1 or 2 edits.
2) The misspelled word is subsequently encoded as a "mask" and input into the fine-tuned model.
3) The model extracts all probable words from the output and matches them against the candidate list.
4) The model then retains a certain number of candidates with the highest contextual scores. Based on our observations, the optimal number is 10.
5) The method proceeds to compare the PERTO similarity between the erroneous word and the remaining replacement candidates. If the error and candidate share the same code, that candidate is considered the most suitable word. However, if two or more probable candidates carry the same PERTO code as the erroneous word, then the candidate with the highest contextual score is selected as the replacement for the error.

Pseudocode3 delivers a comprehensive exploration of the Non-word error correction mechanism.

| | Pseudocode3: Non-Word Error Correction Algorithm |
|---|---|
| **Input:** | Misspelled word $W$, Language Representation Model $LRM$, PERTO algorithm |
| **Output:** | Corrected word $W'$ |
| **Start** | |
| 1 | Generate candidate replacements $R$ using Damerau-Levenshtein edit distance ; |
| 2 | Encode $W$ as a masked word ; |
| 3 | Input $W$ into $LRM$ ; |
| 4 | Extract most probable candidates ($C_k s$) from $LRM$'s output; |
| 5 | Match ($C_k s$ against $R$); |
| 6 | Retain top 10 candidates from $C_k s$ |
| 7 | For each candidate in top 10:<br>    Calculate PERTO similarity score ( $C_k$, $W$);<br>    If PERTO_code($C_k$) == PERTO_code($W$) AND $Probability(C_k) > Probability\ (W)$:<br>        Retain($C_k$) in ($C_k s$)<br>    Else:<br>        Remove ($C_k$) from ($C_k s$) |
| 8 | $W' = \text{MAX}(C_k s)$ |
| 9 | Return $W'$ |
| **End** | |

### 3.7.2 Real-word Error Correction Process

In the scenario of real-word error correction, the process is as follows:
1) The contextual scores of potential candidates are retrieved from the fine-tuned model.
2) The model retains a certain number of candidates with the highest contextual score. Based on our observations, the optimal number is 10.

3) The method then compares the PERTO similarity between the erroneous word and the replacement candidates. If the error and the candidate share the same code, that candidate is deemed the most suitable word.
4) However, if two or more probable candidates carry the same PERTO code as the erroneous word, then the candidate with the highest contextual score is selected as the replacement for the error.

Pseudocode4 delivers a comprehensive exploration of the Non-word error correction mechanism.

Pseudocode4: Real-Word Error Correction Algorithm

| | |
|---|---|
| **Input:** | Misspelled word $W$, Language Representation Model $LRM$, PERTO algorithm |
| **Output:** | Corrected word $W'$ |
| **Start** | |
| 1 | Input $W$ into $LRM$ ; |
| 2 | Extract most probable candidates ($C_k s$) from $LRM$'s output; |
| 3 | Match ($C_k s$ against $R$); |
| 4 | Retain top 10 candidates from $C_k s$ |
| 5 | For each candidate in top 10:<br>    Calculate PERTO similarity score ($C_k, W$);<br>    If PERTO_code($C_k$) == PERTO_code($W$) AND $Probability(C_k) > Probability(W)$:<br>        Retain($C_k$) in ($C_k s$)<br>    Else:<br>        Remove ($C_k$) from ($C_k s$) |
| 7 | $W' = \text{MAX}(C_k s)$ |
| 8 | Return $W'$ |
| **End** | |

## 4. Evaluation and Results

In this section, we first conduct an analysis of the test data. Following this, we evaluate our method's performance and compare it with various baseline models in the task of spelling correction. This comparison will offer valuable insights into the efficacy and precision of our approach in identifying and rectifying spelling errors.

### 4.1 Test Dataset

Our test datasets consist of 188,963 reserved sentences derived from the Persian clinical corpus. Upon scrutinizing the errors present in the test dataset, we found that 1.20% of sentences exhibited instances of non-word errors, which equates to 120 errors in every 10,000 sentences. In addition, 0.29% of sentences contained a real-word error, corresponding to 29 errors in every 10,000 sentences. We examined all the erroneous words to categorize them into one of the predefined classes of errors, such as substitution, transposition, insertion, and deletion. The frequency of these errors, based on the error type, is illustrated in Table 5. When addressing both real-word and non-word errors, substitution errors are more prevalent than other types of errors. Furthermore, insertion errors are quite common when dealing with both classes of error, while deletion and transposition errors are the least common.

We also analyzed the test dataset for the number of edit distances required for spell correction, the results of which are presented in Table 6. In dealing with both real-word and non-word errors, 86.1% of misspellings required an edit distance of 1 to correct the incorrect word. 13.7% of errors were rectified with an edit distance of 2, and a mere 2.1% of errors fell within an edit-distance of 3 or more. Due to the combinatorial explosion when generating and examining candidates within distance 3, these classes of error were excluded from the dataset.

Upon conducting a more thorough analysis of the data, we found that 0.8% of sentences contained more than one error. As our method is designed to handle only one-error-per-sentence, we removed these sentences from the test dataset.

Table 5. Distribution of Different Error Types in the Test Corpus

| Error Type | Substitution | Insertion | Deletion | Transposition |
|---|---|---|---|---|
| Non-word Error | 49.1 | 30.3 | 13.8 | 6.8 |
| Real-word Error | 47.8 | 31.4 | 13.5 | 7.3 |
| **All the Errors** | **48.7** | **30.5** | **13.8** | **7.0** |

Table 6. Minimum Edit Distances Required to Convert Misspelled Words into Correct Words in the Test Dataset

| Edit-Distance | Non-word Error | Real-word Error | Total |
|---|---|---|---|
| 1 | 86.4 | 85.5 | **86.1** |
| 2 | 14.0 | 12.6 | **13.7** |
| 3+ | 2.1 | 1.9 | **2.1** |

### 4.2 Evaluation metrics

The principal metrics for evaluating the effectiveness of models on tasks related to non-word and real-word error identification and rectification are precision (P), recall (R), and the F-measure (F1-Score). Precision (P) quantifies the model's accuracy, whereas recall evaluates its comprehensiveness or sensitivity. The F1-Score, a weighted harmonic average of these two metrics, can be computed by integrating them. In F1, both precision and recall are given equal weight. Equation 4 describes the F1-Score evaluation measure.

$$F1 - Score = 2 * \frac{P * R}{P + R} \quad (4)$$

### 4.3 Baseline Models

In our research, we implemented two baseline models for non-word correction in Persian clinical text to ensure a comprehensive comparison. These models include the four-gram model introduced by [57], and a Persian Continuous Bag-of-Words (CBOW) model [91]. Both models were developed using Python and trained on the same dataset as the pre-trained model. Our aim is to understand the strengths and weaknesses of these models, and leverage this understanding to enhance error correction in Persian language processing. Unfortunately, for real-word error correction in the Persian medical domain, no prior work has been introduced. Therefore, a meaningful comparison is not achievable at this time. This highlights the novelty and importance of our research in this specific area.

#### 4.3.1 Yazdani, et al.

The statistical methodology, pioneered by Yazdani and colleagues, stands out as a promising approach for rectifying non-word errors. It is meticulously crafted to address typographical inaccuracies prevalent in Persian healthcare text, thereby enhancing the quality and reliability of the information [57]. This method leverages a weighted bi-directional fourgram language model to pinpoint the most appropriate substitution for a given error. It incorporates a quadripartite equation that assigns priority to n-grams based on their sequence, thereby enhancing the precision of error correction.

#### 4.3.2 CBOW Model

CBOW model operates by comprehending the semantics of words through the analysis of their surrounding context, and then uses this information as input to predict suitable words for the given context [91]. The architecture of the CBOW model is designed to identify the target word (the center word) based on the context words provided. This model has been specifically trained to tackle the task of non-word error rectification. It employs two matrices to calculate the hidden layer (H): the input matrix (IM) and the output matrix (OM). The CBOW model was trained using a corpus of 1.4 million documents derived from the pre-trained model, which facilitated the generation of the input and output matrices. The training parameters incorporated a context window size of 10 and a dimension size of 300.

### 4.4 Non-word Error Correction Evaluation

In the initial phase of assessment, we juxtapose the effectiveness of our suggested methodology with that of the previously mentioned baseline models concerning non-word error rectification. It's crucial to highlight that all models employ a dictionary look-up method for identifying typos, resulting in an F1-score of 100% for typo detection. Table 7 presents the results of the non-word error correction task, providing a detailed comparison of the effectiveness of our approach and the baseline models.

Table 7. Comparison of Various models' Performance on Non-word Error Correction Task

| Model | Precision | Recall | F1-Score |
|---|---|---|---|
| Pre-trained Model | 88.2 | 89.6 | 88.9 |
| Pre-trained Model + PERTO | **89.3** | **90.7** | **90.0** |
| Yazdani, et al. | 73.8 | 75.4 | 74.6 |
| Continuous Bag-of-Words (CBOW) | 78.7 | 80.8 | 79.7 |

Table 7 provides a detailed comparison of the performance of various models on the non-word error correction task. It compares two configurations of our proposed approach with statistical baselines and the CBOW model. To gauge their effectiveness in practical scenarios, all models were subjected to an extensive array of test instances. The results clearly indicate that both configurations of our approach outperform the other models, demonstrating superior performance. The model achieves its best performance, with an F1-Score of 90.0%, when the PERTO algorithm is employed. The combination of contextual similarity with the PERTO algorithm proves to be the most robust scheme, offering a 1.1% increase in correcting non-word errors compared to using only contextual scores.

The authors of [57] reported achieving an F1-Score of 90.2% for non-word error correction. However, our attempts to reproduce this result in our evaluations were unsuccessful.

In fact, the approach by Yazdani et al. shows the lowest performance, with an F1-Score of 74.6%. The Contextual Scores + PERTO scheme outperforms Yazdani et al.'s approach by 15.4%, further demonstrating the robustness of our method. In terms of the proposed approach, the results of the scheme that combines contextual scores and PERTO are significantly superior to those achieved using only contextual scores. The most effective results are achieved when the pretrained model is used in conjunction with the PERTO orthographic similarity algorithm. Our observations confirm that the PERTO algorithm significantly enhances results, as substitution errors, which are predominantly either visually or phonetically similar, account for 49.1% of all non-word errors in the test corpus. This is in comparison to insertion, deletion, and transposition errors. This underscores the effectiveness of our approach in handling substitution errors.

### 4.5 Real-word Error Detection and Correction Evaluations

We performed a comprehensive evaluation of our proposed model for detecting and correcting real-word errors in Persian clinical text. The results of these evaluations are summarized in Table 8. Our model demonstrated its highest performance in real-word error detection, achieving an F1-Score of 90.6%.

Table 8. Performance Evaluation on Real-word Error Correction Task

| Task | Model | Precision | Recall | F1-Score |
|---|---|---|---|---|
| ***Real-word Error Detection:*** | | | | |

|  | Pre-trained Model | 90.2 | 91.1 | **90.6** |

***Real-word Error Correction:***

|  | Pre-trained Model | 89.4 | 90.7 | 90.0 |
|  | Pre-trained Model + PERTO | 90.8 | 92.2 | **91.5** |

We further evaluated our model's ability to correct real-word errors. As depicted in Table 8, our suggested approach, particularly when enhanced with the PERTO algorithm, exhibits outstanding performance in correcting real-word errors across a range of distances. The model reached its highest F1-Score of 0.915 when the Persian orthographic similarity algorithm was employed, indicating an approximate enhancement of 1.5% in the correction F1-Score. It's noteworthy that the PERTO significantly enhances the results as substitution errors constitute 47.8% of all real-word errors in the test corpus, compared to insertion, deletion, and transposition errors. Furthermore, a significant portion of these substitution errors bear a visual resemblance.

We also conducted a comprehensive analysis of the errors made by our model. We discovered that in a few cases, real-word errors were missed when the erroneous word had a strong semantic connection to the context words. For instance, in the original word sequence "روده اطراف تستیس راست رویت شد" (The presence of the intestine was observed around the right testis.), the medical typist mistakenly replaced the intended word ("روده" / rʊdeh/ 'intestine') with the erroneous word ("توده" /tʊdeh/ 'mass'), which is within an edit distance of 1. This resulted in the word sequence "توده اطراف تستیس راست رویت شد" (A mass was observed surrounding the right testicle), which had a higher context similarity score than the original word sequence. Consequently, this word sequence was overlooked by the model.

While this issue has not been highlighted in previous research on Persian spelling correction, we believe it poses a significant challenge in addressing real-word errors in Persian clinical texts. To prevent such errors from being overlooked, we could present a list of the most probable candidates along with their context scores to a human expert, allowing them to select the most appropriate replacement. This emphasizes that, despite the advancements in state-of-the-art models, human expertise remains indispensable in certain situations.

In summary, the results indicate that our proposed method exhibits robustness and precision in detecting and rectifying context-sensitive errors in Persian clinical text, thereby affirming its potential for practical application in the field.

## 5. Discussion

Typographical errors, a frequent occurrence in radiology reports often attributed to incessant interruptions and a dynamic work environment, have the potential to endanger patient health, introduce ambiguity, and undermine the reputation of radiologists [92]. The cardinal goal of our research was to pioneer an avant-garde technique for pinpointing and rectifying spelling inaccuracies in Persian clinical text. The elaborate morphology and syntax of the Persian language, intertwined with the pivotal role of meticulous documentation in fostering effective patient care, facilitating research, and safeguarding patient safety, accentuate the gravity of this undertaking.

Within the confines of the Imaging Department at Imam Khomeini Hospital, the formulation of radiology reports is an intricate multi-step endeavor that averages around 30 minutes in duration.

This process includes dictation by radiologists, transcription by medical typists, and a review and editing process before the final report is stored in the HIS. However, this process includes non-value-added activities, known as 'Muda', particularly the time spent between transcription and final confirmation [93]. Our newly developed software addresses this inefficiency by quickly correcting misspelled words during transcription, reducing the time between initial writing and final confirmation, and thereby decreasing 'Muda'.

Our approach leverages a pre-trained language representation model, fine-tuned specifically for the task of spelling correction in the clinical domain. This model is complemented by an innovative orthographic similarity matching algorithm, PERTO, which uses visual similarity of characters for ranking correction candidates. This unique combination of techniques distinguishes our approach from existing methods, enabling our model to effectively address both non-word and real-word errors. The evaluation of our approach demonstrated its robustness and precision in detecting and rectifying word errors in Persian clinical text. In terms of non-word error correction, our model achieved an F1-Score of 90.0% when the PERTO algorithm was employed. This represents a 1.1% increase in correcting non-word errors compared to using only contextual scores. For real-word error detection, our model demonstrated its highest performance, achieving an F1-Score of 90.6%. Furthermore, the model reached its highest F1-Score of 91.5% for real-word error correction when the PERTO algorithm was employed, indicating an approximate enhancement of 1.5% in the correction F1-Score.

Despite these promising results, our model has certain limitations. For instance, in a few cases, real-word errors were missed when the erroneous word had a strong semantic connection to the context words. Additionally, while our model is effective in handling non-word and real-word errors, it is not equipped to deal with grammatical errors. Moreover, our model was set up to handle one-error-per-sentence cases and cannot handle more than one error in a sentence. There were a few cases where a sentence included more than two errors.

Building upon our current achievements, the integration of emerging technologies such as BCI eye-tracking, VR/AR, and EEG offers a promising frontier for further enhancing our spelling correction system. These technologies present unique opportunities to address some of the inherent limitations identified in our study. For example, BCIs could offer intuitive, direct error correction interfaces, while eye-tracking might refine error detection based on user interaction patterns. VR/AR could provide immersive training environments, improving proficiency with correction tools, and EEG monitoring could lead to spelling correction interfaces that adapt to user stress levels and cognitive states, ultimately making the correction process less taxing and more efficient.

While prevailing spelling correction mechanisms for the Persian language cater to a broad spectrum and are not tailored to the medical sphere, our innovative system is specifically architected to autonomously pinpoint and amend misspellings prevalent in Persian radiology and ultrasound reports. The seamless integration of automatic spell-checking systems, notably in critical facets for patient safety such as allergy entries, medication details, diagnoses, and problem listings, can substantially bolster the quality and exactness of electronic medical records. Our system, which can be seamlessly integrated as an auxiliary program on platforms like Microsoft Office Word, web-browsers, or employed as an API in the HIS system, expands the potential applications of our model transcending the boundaries of the clinical domain.

In summary, the results of this study affirm the potential of our proposed method in transforming Persian clinical text processing. By effectively addressing the unique challenges posed by the Persian language and integrating cutting-edge technologies, our approach paves the way for more accurate and efficient clinical documentation, contributing to improved patient care and safety.

## 5. Conclusions

This study presents a novel method for detecting and correcting spelling errors in Persian clinical texts, leveraging a pre-trained model fine-tuned for this specific domain. Our approach has notably outperformed existing models, achieving F1-Scores of over 90% in both real-word and non-word error correction. This advancement underscores the method's robustness and its wide-ranging applicability, from error types like substitution and insertion to deletion and transposition. By integrating our orthographic similarity algorithm, PERTO, with contextual insights, we've significantly enhanced the correction success rate, marking a substantial improvement in spelling error correction for Persian clinical texts.

The potential of our methodologies extends beyond medical documentation, offering valuable applications in engineering sciences. The NLP and machine learning techniques employed here could revolutionize error detection and correction in engineering documents and software code, improving review processes, technical documentation accuracy, and software development efficiency. Furthermore, our findings could inform the creation of intelligent diagnostic systems for predictive maintenance and quality control, leveraging our error correction mechanisms for enhanced precision and reliability.

Looking ahead, we aim to refine our model further to tackle multiple errors within a sentence and address grammatical inaccuracies, broadening our method's comprehensiveness for the Persian medical domain. Additionally, we plan to explore the integration of emerging technologies like BCI, eye-tracking, VR/AR, and EEG, aiming to create more intuitive correction interfaces and immersive training environments. These efforts will not only advance spelling correction tools technically but also amplify their practical impact in medical documentation, contributing to improved patient care and safety.

## Declarations

### Ethics approval and consent to participate
The study was conducted in accordance with ethical standards and received approval from the Institutional Review Board of the Islamic Azad University, Kerman Branch (approval ID: IR.IAU.KERMAN.REC.1402.124). As the study did not involve any human trials, the requirement for informed consent was waived by the same Institutional Review Board.

### Consent for Publication
Not applicable.

### Availability of Data and Materials
The data that support the findings of this study are held by Imam Khomeini Hospital. They are not publicly accessible due to privacy restrictions. However, they may be available from the authors upon reasonable request and with permission from the hospital.

### Competing Interests
The authors declare that they have no competing interests concerning the publication of this paper.

## Funding

This research did not receive any specific grant from funding agencies in the public, commercial, or not-for-profit sectors.

## Authors' Contributions

Seyed Mohammad Sadegh Dashti and Seyedeh Fatemeh Dashti conceptualized and designed the study. Seyed Mohammad Sadegh Dashti developed the model and performed the experiments. Seyedeh Fatemeh Dashti collected and analyzed the data. Both authors contributed to writing the manuscript and approved the final version for publication.

29